\documentclass{article}

\usepackage[preprint]{sty}

\usepackage[utf8]{inputenc} % allow utf-8 input
\usepackage[T1]{fontenc}    % use 8-bit T1 fonts
\usepackage{hyperref}       % hyperlinks
\usepackage{url}            % simple URL typesetting
\usepackage{booktabs}       % professional-quality tables
\usepackage{amsfonts}       % blackboard math symbols
\usepackage{nicefrac}       % compact symbols for 1/2, etc.
\usepackage{microtype}      % microtypography
\usepackage{xcolor}         % colors
\usepackage{amsmath}
\usepackage{algorithm}
\usepackage{algpseudocode}
\usepackage{graphicx}
\usepackage{multirow} 
\usepackage[most]{tcolorbox}
\usepackage{fancyvrb}
\usepackage{array}
\usepackage[table]{xcolor}
\usepackage{makecell}

\definecolor{codeColor}{HTML}{4F378B}   % Code      (purple)
\definecolor{ctrlColor}{HTML}{2E7D32}   % Control   (green, placeholder)
\definecolor{knowColor}{HTML}{E65100}   % Knowledge (orange, placeholder)
\definecolor{semColor} {HTML}{1565C0}   % Semantic  (blue,   placeholder)

\title{VulTriage: Triple-Path Context Augmentation for LLM-Based Vulnerability Detection}

\author{%
\begin{tabular}{c}
Wenxin Tang\textsuperscript{1} \quad
Xiang Zhang\textsuperscript{2} \quad
Junliang Liu\textsuperscript{3} \quad
Jingyu Xiao\textsuperscript{4} \quad
Xi Xiao\textsuperscript{1} \\
Jinlong Yang\textsuperscript{5} \quad
Yuehe Ma\textsuperscript{6} \quad
Zhenyu Liu\textsuperscript{5} \quad
Zhengheng Li\textsuperscript{7} \quad
Zicheng Wang\textsuperscript{8} \\
Wang Luo\textsuperscript{9} \quad
Qing Li\textsuperscript{10} \quad
Lei Wang\textsuperscript{11} \quad
Peng Xiangli\textsuperscript{12} \\
\\
\textsuperscript{1}Tsinghua University \quad
\textsuperscript{2}Henan University \quad
\textsuperscript{3}Dalian Maritime University \\
\textsuperscript{4}The Chinese University of Hong Kong \quad
\textsuperscript{5}Northwestern Polytechnical University \\
\textsuperscript{6}BNU-HKBU United International College \quad
\textsuperscript{7}Southeast University \\
\textsuperscript{8}Jilin University \quad
\textsuperscript{9}Sun Yat-sen University \quad
\textsuperscript{10}Peng Cheng Laboratory \\
\textsuperscript{11}Guangzhou Intelligence Communications Technology Co., Ltd. \\
\textsuperscript{12}The Fifth Electronic Research Institute of MIIT \\
\\
\texttt{twx24@mails.tsinghua.edu.cn} \quad
\texttt{gnaix@henu.edu.cn} \quad
\texttt{xiaox@sz.tsinghua.edu.cn}
\end{tabular}
}

\begin{document}

\maketitle

\begin{abstract}
Automated vulnerability detection is a fundamental task in software security, yet existing learning-based methods still struggle to capture the structural dependencies, domain-specific vulnerability knowledge, and complex program semantics required for accurate detection. Recent Large Language Models (LLMs) have shown strong code understanding ability, but directly prompting them with raw source code often leads to missed vulnerabilities or false alarms, especially when vulnerable and benign functions differ only in subtle semantic details. To address this, we propose VulTriage, a triple-path context augmentation framework for LLM-based vulnerability detection. VulTriage enhances the LLM input through three complementary paths: a Control Path that extracts and verbalizes AST, CFG, and DFG information to expose control and data dependencies; a Knowledge Path that retrieves relevant CWE-derived vulnerability patterns and examples through hybrid dense--sparse retrieval; and a Semantic Path that summarizes the functional behavior of the code before the final judgment. These contexts are integrated into a unified instruction to guide the LLM toward more reliable vulnerability reasoning. Experiments on the PrimeVul pair test set show that VulTriage achieves state-of-the-art performance, outperforming existing deep learning and LLM-based baselines on key pair-wise and classification metrics. Further ablation studies verify the effectiveness of each path, and additional experiments on the Kotlin dataset demonstrate the generalization ability of VulTriage under low-resource and class-imbalanced settings. Our
code is available at ~\url{https://github.com/vinsontang1/VulTriage}
\end{abstract}

\section{Introduction}
\label{sec:introduction}

Software vulnerabilities remain a fundamental threat to modern software systems, and automated vulnerability detection has long been an important research topic in both academia and industry. Existing deep learning-based methods, including GNN-based models and pre-trained code models, have made substantial progress~\cite{zhou2019devign,chakraborty2021deep,feng2020codebert}. However, these methods still face practical limitations: GNN-based methods often require complex graph construction and high-quality labels~\cite{croft2023data}, while pre-trained code models usually treat source code as a linear token sequence and struggle to capture structural dependencies and execution semantics~\cite{fu2022linevul,zhang2023vulnerability}.

Recent advances in Large Language Models (LLMs) have shown strong capabilities across diverse domains, such as visual understanding~\cite{liu2025benchmarking,xiao2025designbench,luo2026rethinking}, code generation~\cite{tang2025slidecoder,tang2026efficientpostergen,xiao2025interaction2code}, and security analysis~\cite{zou2025queryattack}. Building on this progress, LLM-driven vulnerability detection has become a promising direction. Recent methods such as VulTrial~\cite{widyasari2025let} and ConColl~\cite{tsai2025sequential} further improve vulnerability detection through multi-agent reasoning, and staged decision making. Despite their effectiveness, directly relying on LLMs to judge raw source code still leaves several important challenges unaddressed.

\textbf{First, LLMs suffer from structural blindness.} Source code is usually processed as a token sequence, while many vulnerabilities are determined by cross-statement control-flow paths and data-propagation chains. For example, a buffer overflow may depend on whether user-controlled data reaches a sensitive write operation without proper boundary checks. Such structural evidence is difficult to recover from raw tokens alone.

\textbf{Second, LLMs face knowledge deficiency.} The security knowledge acquired during general-purpose pre-training is implicit and scattered, whereas vulnerability detection often requires explicit domain priors, such as vulnerability taxonomies, definitions, and vulnerable code examples. The Common Weakness Enumeration (CWE) database~\cite{mitre_cwe} provides a systematic collection of such knowledge, but it is not explicitly grounded in ordinary LLM prompting.

\textbf{Third, LLMs still have a semantic gap when analyzing complex program logic.} Real-world vulnerable code often involves pointer operations, nested branches, implicit type conversions, and subtle boundary checks. Even a small semantic misunderstanding, such as overlooking a check or misinterpreting a data dependency, can reverse the final vulnerability judgment.

To address the aforementioned limitations, we introduce \textbf{VulTriage}, an LLM-based vulnerability detection framework built upon Triple-Path Context Augmentation. The core idea is to avoid asking the LLM to directly judge raw source code; instead, VulTriage first constructs three complementary contexts for the final decision. \textbf{First, the Control Path extracts AST, CFG, and Data-Flow Graph (DFG) information and converts them into compact natural-language structural descriptions. Second, the Knowledge Path retrieves relevant vulnerability descriptions and vulnerable code examples from the CWE knowledge base through hybrid dense--sparse retrieval~\cite{lewis2020retrieval,chen2024bge}. Third, the Semantic Path summarizes the functional behavior of the input code, providing a denoised semantic view before the final judgment.} These contexts are integrated into a unified prompt to guide the LLM toward more reliable vulnerability reasoning.

Our contributions are summarized as follows:
\begin{itemize}
    \item We propose VulTriage, an LLM-based vulnerability detection framework driven by Triple-Path Context Augmentation, which integrates structural context, vulnerability knowledge, and semantic explanation into a unified decision context for vulnerability reasoning.

    \item We design a Control Path that compresses AST, CFG, and DFG information into LLM-friendly natural-language descriptions, enabling LLMs to better perceive vulnerability-relevant control and data dependencies.

    \item We introduce a Knowledge Path based on hybrid dense--sparse retrieval, which dynamically retrieves CWE-derived vulnerability descriptions and examples to provide explicit domain priors.

    \item We conduct extensive experiments on PrimeVul and additional analyses on ablation settings, low-resource Kotlin data, and attention-based context utilization. The results demonstrate the effectiveness and generalization ability of VulTriage.
\end{itemize}

\section{Related Work}

\subsection{DL-Based Vulnerability Detection}

DL-based vulnerability detection methods mainly rely on semantic or structural code representations. Semantic methods treat source code as token sequences and use pre-trained code models such as CodeBERT~\cite{feng2020codebert}, CodeT5~\cite{wang2021codet5}, UniXCoder~\cite{guo2022unixcoder}, and LineVul~\cite{fu2022linevul} for vulnerability prediction. Some later studies further enhance semantic representations through comments~\cite{wen2024scale}, identifier perturbation, counterfactual augmentation~\cite{kuang2023leveraging}, or noisy-label learning~\cite{wen2023less}. Structure-aware methods instead introduce ASTs, CFGs, DFGs, PDGs, slices~\cite{zhou2019devign,jiang2024dfept,liu2024pre,weng2024matsvd}, or execution paths~\cite{zhang2023vulnerability,peng2023ptlvd} to capture syntactic, control-flow, and data-flow dependencies, often using GNNs, hybrid Transformer--GNN models~\cite{wang2024combining}, or dependency-aware attention mechanisms. However, these methods usually require task-specific training, complex graph construction, or customized architectures. VulTriage differs by converting structural information into compact natural-language contexts that can directly support LLM reasoning.

\subsection{LLM-Based Vulnerability Detection}

Recent work has explored LLMs for vulnerability detection through direct prompting, chain-of-thought reasoning~\cite{ding2024primevul}, multi-agent collaboration~\cite{widyasari2025let,tsai2025sequential}, or retrieval-augmented generation. These methods improve flexibility and interpretability, especially when external vulnerability knowledge such as CWE~\cite{mitre_cwe} descriptions and examples is introduced. Nevertheless, directly prompting LLMs with raw code can be unstable, since the model must simultaneously understand program semantics, trace dependencies, and recall security knowledge. Existing knowledge-augmented or multi-agent methods also often lack explicit structural evidence or introduce additional inference cost. In contrast, VulTriage integrates structural context, retrieved vulnerability knowledge, and semantic explanation into a unified prompt, enabling more reliable LLM-based vulnerability detection.

\begin{figure*}
    \centering
    \includegraphics[width=0.95\linewidth]{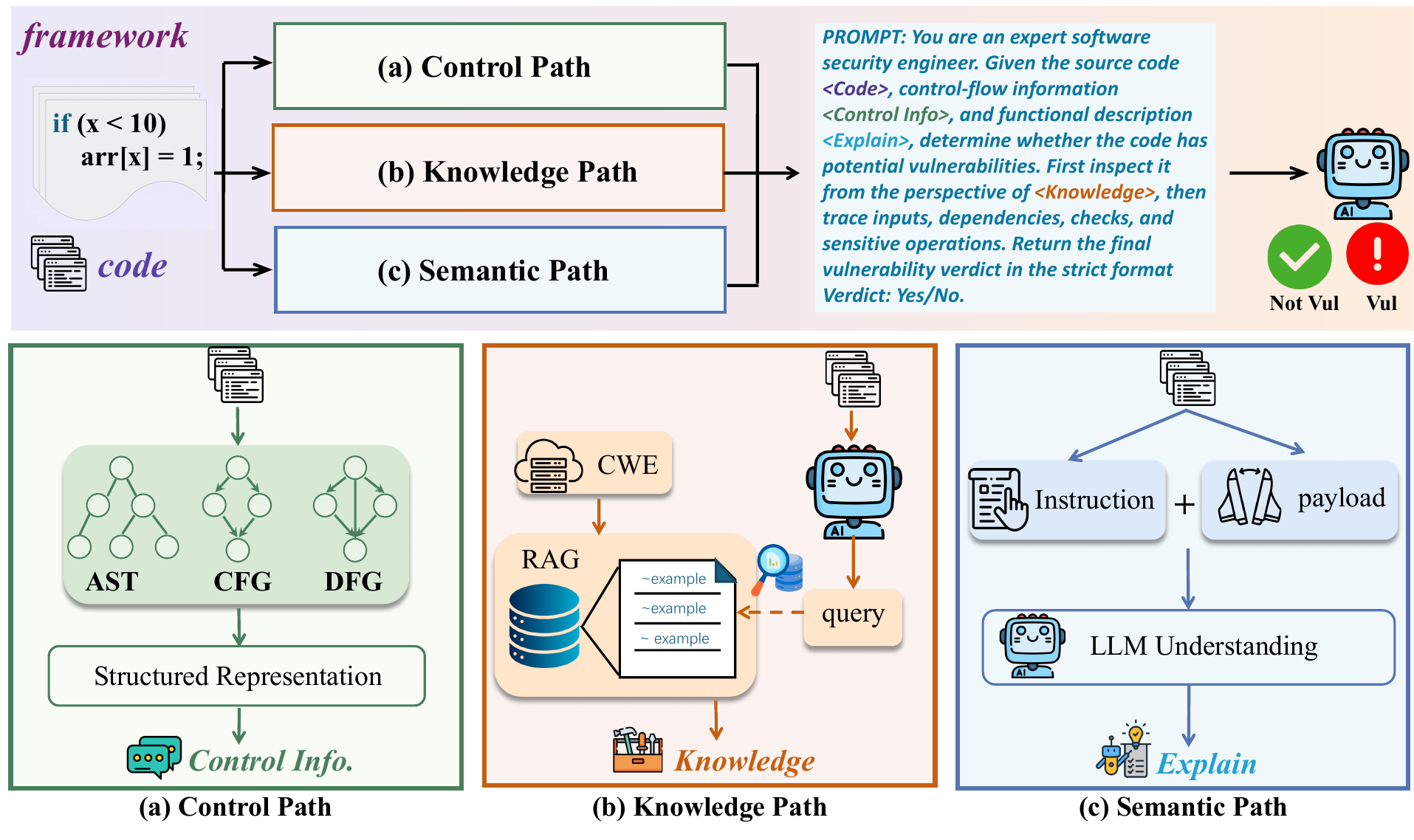}
    \caption{The framework of VulTriage.}
    \label{fig:framework}
\end{figure*}

\section{Methodology}
\label{sec:methodology}

In this section, we introduce VulTriage, a framework that augments LLM-driven vulnerability detection with three complementary context streams. Given a piece of source code $C$ as input, VulTriage predicts a binary label $\hat{y} \in \{0, 1\}$ indicating whether $C$ contains a vulnerability. VulTriage comprises three core modules. First, a \textbf{Control Path} parses $C$ into three structural graphs and produces a natural-language structural context $\mathbf{S}$ that exposes control and data dependencies to the LLM. Second, a \textbf{Knowledge Path} retrieves vulnerability patterns from a CWE-derived knowledge base through a hybrid dense--sparse retrieval mechanism, producing a knowledge context $\mathbf{K}$. Third, a \textbf{Semantic Path} queries the LLM itself to obtain a functional-level description of $C$, producing a semantic context $\mathbf{E}$. The three contexts together with $C$ are assembled into a unified instruction $\mathcal{T}$, and the final verdict is produced as
\begin{equation}
    \hat{y} \;=\; \mathcal{M}_{\text{LLM}}\!\bigl(\mathcal{T}(C, \mathbf{S}, \mathbf{K}, \mathbf{E})\bigr),
\end{equation}
where $\mathcal{M}_{\text{LLM}}$ denotes the LLM used for the final judgment. The overall framework is illustrated in Figure~\ref{fig:framework}, and the structure of $\mathcal{T}$ is detailed in \S\ref{sec:semantic-path}.

% ====================================================================
\subsection{Control Path}
\label{sec:control-path}

To expose the static structure of source code to the LLM and alleviate its structural blindness on control and data dependencies, we propose the Control Path, a three-stage pipeline that distills the source code into a compact natural-language structural context $\mathbf{S}$. The pipeline consists of graph construction, multi-granularity filtering, and structured representation generation.

\subsubsection{Graph Construction}
\label{sec:graph-construction}

The Control Path first parses the source code $C$ into three complementary structural representations. The Abstract Syntax Tree (AST) captures the syntactic hierarchy and statement composition, the Control-Flow Graph (CFG) captures the execution order and branching among statements, and the Data-Flow Graph (DFG) captures the def--use propagation of variables across statements. The three graphs characterize the code along three orthogonal dimensions---syntax, control, and data---and together provide a complete structural basis for the subsequent vulnerability-relevant information extraction.

\subsubsection{Multi-Granularity Filtering}
\label{sec:multi-granularity}

Directly feeding the three graphs to the LLM is problematic: the raw graphs contain a large number of semantically irrelevant nodes (e.g., type-expansion nodes in the AST and purely sequential statement fragments in the CFG) that dilute vulnerability-related signals, and their total size typically exceeds the effective context window of the LLM. To mitigate this, the Control Path introduces a multi-granularity filtering mechanism parameterized by a granularity level $\ell \in \{A, B, C\}$. At level $A$, only skeleton nodes are retained, including function definitions, branches, loops, function calls, and returns, which yields the most concise view of the code. Level $B$ extends level $A$ by additionally retaining assignments, declarations, and key operators, offering a balanced view suitable for most vulnerability patterns. Level $C$ retains all semantic nodes while only removing noise nodes such as type expansions, preserving fine-grained details at the cost of a larger representation. The path-enumeration upper bound of the CFG and the data-chain length upper bound of the DFG are scaled accordingly with $\ell$, which together control the overall size of the resulting natural-language description.

\subsubsection{Structured Representation Generation}
\label{sec:structured-repr}

\begin{algorithm}[t]
\caption{Structured Representation Generation}
\label{alg:control}
\begin{algorithmic}[1]
    \Require source code $C$; granularity level $\ell \in \{A, B, C\}$
    \Ensure structural context $\mathbf{S}$
    \State $(G_{\text{ast}},\, G_{\text{cfg}},\, G_{\text{dfg}}) \leftarrow \mathrm{Parse}(C)$
    \Statex \textit{/* Multi-granularity filtering */}
    \State $G'_{\text{ast}} \leftarrow \Pi^{\text{ast}}_{\ell}(G_{\text{ast}})$
    \State $G'_{\text{cfg}} \leftarrow \Pi^{\text{cfg}}_{\ell}(G_{\text{cfg}})$
    \State $G'_{\text{dfg}} \leftarrow \Pi^{\text{dfg}}_{\ell}(G_{\text{dfg}})$
    \Statex \textit{/* Salient-view extraction with budget $b(\ell)$ */}
    \State $\mathcal{V}_{\text{ast}} \leftarrow \bigl\{\,\mathrm{Aggr}(f) \,\bigm|\, f \in \mathcal{F}(G'_{\text{ast}})\,\bigr\}$
    \State $\mathcal{V}_{\text{cfg}} \leftarrow \mathrm{Paths}_{b(\ell)}\!\bigl(G'_{\text{cfg}}\bigr)$
    \State $\mathcal{V}_{\text{dfg}} \leftarrow \mathrm{Chains}_{b(\ell)}\!\bigl(G'_{\text{dfg}}\bigr)$
    \Statex \textit{/* Verbalization via a shared operator $\tau$ */}
    \State $T_{\text{ast}} \leftarrow \tau(\mathcal{V}_{\text{ast}});\;\;
            T_{\text{cfg}} \leftarrow \tau(\mathcal{V}_{\text{cfg}});\;\;
            T_{\text{dfg}} \leftarrow \tau(\mathcal{V}_{\text{dfg}})$
    \State $\mathbf{S} \leftarrow T_{\text{ast}} \oplus T_{\text{cfg}} \oplus T_{\text{dfg}}$
    \State \Return $\mathbf{S}$
\end{algorithmic}
\end{algorithm}

As shown in Algorithm~\ref{alg:control}, the Control Path starts by parsing the source code $C$ once and jointly producing the three structural representations $G_{\text{ast}}$, $G_{\text{cfg}}$, and $G_{\text{dfg}}$ (line~1). To suppress noise and keep the graphs within a tractable budget, granularity-dependent filters $\Pi^{\cdot}_{\ell}$ are applied to each graph (lines~2--4): type-expansion nodes such as \texttt{TypeDecl} and \texttt{PtrDecl} are removed from the AST while function definitions, declarations, assignments, branches, and calls are preserved~\cite{pycparser}; purely sequential fragments in the CFG are folded into single edges while entry/exit, branches, loops, and call nodes are retained; the DFG keeps parameter-source nodes together with cross-statement def--use edges.

On top of the filtered graphs, the Control Path extracts \emph{salient views} under a shared budget $b(\ell)$ that controls the output scale (lines~5--7). Concretely, $\mathrm{Aggr}$ aggregates each function $f \in \mathcal{F}(G'_{\text{ast}})$ in the AST into a function-level semantic summary, $\mathrm{Paths}_{b(\ell)}$ performs bounded path enumeration on the CFG from the entry to the exit and prioritizes skeleton paths that carry branches and call sites, and $\mathrm{Chains}_{b(\ell)}$ traces def--use chains starting from each parameter node in the DFG until a call or return sink is reached. The three sets of salient views correspond to the three categories of information essential to vulnerability analysis, namely code composition, execution paths, and data propagation.

The salient views are then rendered into natural-language fragments through a shared verbalization operator $\tau$, which is instantiated by a small set of pre-defined templates that fill graph elements into natural-language skeletons (line~8). The resulting fragments are concatenated in a fixed order to form the final structural context $\mathbf{S}$ (line~9). Concrete templates used by $\tau$ and a full instantiation example are provided in Appendix~\ref{sec:appendix-templates}.

% ====================================================================
\subsection{Knowledge Path}
\label{sec:knowledge-path}

To compensate for the implicit and scattered vulnerability knowledge learned by the LLM during general-purpose pre-training, we propose the Knowledge Path, which injects explicit domain priors into the LLM at inference time through a retrieval-augmented mechanism. The Knowledge Path operates in three steps: query generation, hybrid dense--sparse retrieval, and knowledge assembly.

\subsubsection{Knowledge Base Construction}
\label{sec:kb-construction}

We build the retrieval corpus directly from the Common Weakness Enumeration (CWE) database, which is a community-maintained taxonomy covering hundreds of vulnerability types~\cite{mitre_cwe}. For each CWE entry, its name, description, and vulnerable code example are extracted and concatenated into a single indexable passage, yielding a retrieval corpus $\mathcal{D} = \{d_i\}_{i=1}^{N}$.

\subsubsection{Query Generation}
\label{sec:query-generation}

Using the raw source code $C$ as the retrieval query causes a severe modality mismatch against the natural-language descriptions of CWE entries. To close this gap, the Knowledge Path first issues an instruction to the LLM that requests a coarse prediction of the potential vulnerability types of $C$, with the output constrained to at most two natural-language vulnerability descriptions $q_1$ and $q_2$. These descriptions are then used as retrieval queries. When the LLM judges $C$ to be free of vulnerabilities, a fallback query is issued to retrieve a generic vulnerability pattern for further inspection.

\subsubsection{Hybrid Dense--Sparse Retrieval}
\label{sec:hybrid-retrieval}

Matching each query $q$ to the knowledge corpus $\mathcal{D}$ requires both semantic generalization, so that queries phrased differently from a CWE entry can still be matched, and lexical precision, so that security-critical terms such as ``buffer overflow'' and ``null pointer'' are not dropped by a purely semantic matcher. To meet both requirements, the Knowledge Path employs BGE-M3~\cite{chen2024bge} to produce two complementary representations for every text: a dense vector $\mathbf{e}^{d}(\cdot)$ that encodes holistic semantics, and a sparse lexical weight vector $\mathbf{e}^{s}(\cdot)$ that encodes term-level importance. The similarity between a query $q$ and an entry $d_i$ is then defined as a convex combination of the two components,
\begin{equation}
    \mathrm{sim}(q, d_i) \;=\; \alpha \cdot \cos\!\bigl(\mathbf{e}^{d}(q),\, \mathbf{e}^{d}(d_i)\bigr) \;+\; (1 - \alpha) \sum_{t \in q \cap d_i} \mathbf{e}^{s}(q)[t] \cdot \mathbf{e}^{s}(d_i)[t],
\end{equation}
where the first term is the cosine similarity between the dense vectors, the second term is a dot product over co-occurring terms $t$ weighted by the sparse encoders, and $\alpha \in [0, 1]$ is the fusion coefficient balancing semantic generalization and term-level precision. For each query $q_j$, the top-$K$ most similar CWE entries are returned as the retrieval result.

\subsubsection{Knowledge Assembly}
\label{sec:knowledge-assembly}

All retrieved entries are deduplicated and concatenated in order into a single natural-language passage, in which each entry carries the vulnerability name, description, and vulnerable code example. The resulting passage is the knowledge context $\mathbf{K}$, which anchors the LLM's subsequent judgment to relevant vulnerability patterns.

% ====================================================================
\subsection{Semantic Path and Instruction Construction}
\label{sec:semantic-path}

\subsubsection{Semantic Path}
\label{sec:semantic}

While the Control Path provides a structural view of the code and the Knowledge Path supplies external vulnerability anchors, the LLM may still misinterpret the functional intent of the code when confronted with multi-level pointer dereferencing, deeply nested branches, or implicit type conversions. To mitigate this semantic gap, we introduce the Semantic Path, which obtains a controlled functional-level description of $C$ before the final judgment and thus serves as a denoised view of the raw code.

Concretely, the Semantic Path queries the LLM to analyze the source code $C$ and identify key operations relevant to program behavior, including input handling, memory operations, file operations, and system calls. Beyond these operations, the LLM is also asked to attend to auxiliary signals that reflect programmer intent, such as variable names, function names, and inline comments, and to describe how data flows through the program when observable. The output of this stage is a concise natural-language explanation $\mathbf{E}$. The instruction further restricts the LLM to describe only directly observable behaviors, avoiding speculation about unobservable intent.

\subsubsection{Instruction Construction}
\label{sec:instruction}

The three context streams $(\mathbf{S}, \mathbf{K}, \mathbf{E})$ and the original code $C$ are assembled into a unified instruction $\mathcal{T}$ that serves as the final input to the LLM. As shown in Figure~\ref{fig:framework}, $\mathcal{T}$ consists of four parts: a role specification that sets the LLM as a static code auditor, an input description that introduces the four input slots, a set of internal analysis rules that guide the LLM's reasoning, and a strict output format that restricts the response to a Yes/No verdict. The four input slots are filled with the corresponding components: \textcolor{codeColor}{\textbf{\textit{<Code>}}} is the original source code $C$, \textcolor{ctrlColor}{\textbf{\textit{<Control Info>}}} is the structural context $\mathbf{S}$ produced by the Control Path, \textcolor{knowColor}{\textbf{\textit{<Knowledge>}}} is the knowledge context $\mathbf{K}$ produced by the Knowledge Path, and \textcolor{semColor}{\textbf{\textit{<Explain>}}} is the semantic context $\mathbf{E}$ produced by the Semantic Path. The LLM consumes the assembled instruction and produces the final vulnerability verdict $\hat{y}$.

% ===================================================================
%  Required preamble (add once if not already present):
%    \usepackage{booktabs}
%    \usepackage{graphicx}
%    \usepackage[table]{xcolor}
% ===================================================================

\begin{table*}[t]
\centering
\caption{Performance comparison on the PrimeVul pair test set. $\uparrow$ indicates higher is better; $\downarrow$ indicates lower is better. Best results on the primary metrics (P-C, Error, Acc, and F1) are in \textbf{bold}. Superscripts: $\dagger$ fine-tuned on the full PrimeVul train/val set; $\ddagger$ fine-tuned on the paired PrimeVul train/val subset; $\S$ released checkpoint from LineVul; $*$ best-performing prompting strategy reported in PrimeVul.}
\label{tab:main}
\renewcommand{\arraystretch}{1.15}
\resizebox{0.95\textwidth}{!}{%
\begin{tabular}{@{}l
                 >{\centering\arraybackslash}p{0.8cm}
                 >{\centering\arraybackslash}p{0.8cm}
                 >{\centering\arraybackslash}p{0.8cm}
                 >{\centering\arraybackslash}p{0.8cm}
                 >{\centering\arraybackslash}p{0.95cm}
                 >{\centering\arraybackslash}p{0.8cm}
                 >{\centering\arraybackslash}p{0.8cm}
                 >{\centering\arraybackslash}p{0.9cm}
                 >{\centering\arraybackslash}p{0.95cm}
                 >{\centering\arraybackslash}p{0.95cm} @{}}
\toprule
\textbf{Method}
 & \textbf{P-C$\uparrow$} & \textbf{P-V$\downarrow$} & \textbf{P-B$\downarrow$} & \textbf{P-R$\downarrow$} & \textbf{Error$\downarrow$}
 & \textbf{P$\uparrow$} & \textbf{R$\uparrow$} & \textbf{FPR$\downarrow$}
 & \textbf{Acc$\uparrow$} & \textbf{F1$\uparrow$} \\
\midrule
%---------- Section 1: DL-based methods ----------
\multicolumn{11}{@{}l}{\textit{DL-based methods}} \\
\addlinespace[2pt]
CodeBERT$^{\dagger}$
 & 5  & 37  & 386 & 7  & 430 & 0.4884 & 0.0966 & 0.1011 & 0.4977 & 0.1612 \\
CodeBERT$^{\ddagger}$
 & 0  & 435 & 0   & 0  & 435 & 0.5000 & 1.0000 & 1.0000 & 0.5000 & 0.6667 \\
CodeT5$^{\dagger}$
 & 0  & 57  & 373 & 5  & 435 & 0.4790 & 0.1310 & 0.1425 & 0.4943 & 0.2058 \\
CodeT5$^{\ddagger}$
 & 0  & 434 & 1   & 0  & 435 & 0.5000 & 0.9977 & 0.9977 & 0.5000 & 0.6662 \\
UniXCoder$^{\dagger}$
 & 4  & 26  & 403 & 2  & 431 & 0.5172 & 0.0690 & 0.0644 & 0.5023 & 0.1217 \\
UniXCoder$^{\ddagger}$
 & 6  & 412 & 17  & 0  & 429 & 0.5036 & 0.9609 & 0.9471 & 0.5069 & 0.6609 \\
LineVul$^{\S}$
 & 7  & 89  & 336 & 3  & 428 & 0.5106 & 0.2207 & 0.2115 & 0.5046 & 0.3082 \\
LineVul$^{\ddagger}$
 & 0  & 434 & 1   & 0  & 435 & 0.5000 & 0.9977 & 0.9977 & 0.5000 & 0.6662 \\
\midrule
%---------- Section 2: LLM-based methods ----------
\multicolumn{11}{@{}l}{\textit{LLM-based methods (GPT-4o)}} \\
\addlinespace[2pt]
Ding et al.'s CoT$^{*}$
 & 40 & 43  & 323 & 29 & 395 & 0.5355 & 0.1908 & 0.1655 & 0.5126 & 0.2814 \\
GPTLens
 & 44 & 241 & 122 & 28 & 391 & 0.5144 & 0.6552 & 0.6184 & 0.5184 & 0.5763 \\
VulTrial
 & 81 & 179 & 125 & 50 & 354 & 0.5317 & 0.5977 & 0.5264 & 0.5356 & 0.5628 \\
ConColl
 & 22 & 337 & 65  & 11 & 413 & 0.5078 & 0.8253 & 0.8000 & 0.5126 & 0.6287 \\
\rowcolor{gray!12}
\textbf{VulTriage (Ours)}
 & \textbf{147} & 207 & 52 & 29 & \textbf{288}
 & 0.6000 & 0.8138 & 0.5425
 & \textbf{0.6356} & \textbf{0.6907} \\
\bottomrule
\end{tabular}%
}
\end{table*}

\section{Experiments}
\label{sec:experiments}

\subsection{Experimental Setup}
\label{sec:exp-setup}

\noindent\textbf{Dataset.}
We evaluate VulTriage on PrimeVul~\cite{ding2024primevul}, a recently released function-level vulnerability detection benchmark. Following VulTrial~\cite{widyasari2025let}, we adopt the PrimeVul \emph{pair} split, in which each vulnerable function is coupled with its fixed counterpart such that the two share at least 80\% of their string content, forcing a model to discriminate based on subtle semantic differences rather than surface statistics. The pair split provides 3{,}789 training pairs, 480 validation pairs, and 435 test pairs, with no data leakage across splits. All results are reported on the 435 test pairs (870 functions).

\noindent\textbf{Baselines.}
We compare VulTriage against two groups of baselines: deep learning-based methods (DL-based) and LLM-based methods.

For the DL-based methods, we select four pre-trained code models widely adopted in prior vulnerability detection literature: CodeBERT~\cite{feng2020codebert}, CodeT5~\cite{wang2021codet5}, UniXCoder~\cite{guo2022unixcoder}, and LineVul~\cite{fu2022linevul}. Following VulTrial~\cite{widyasari2025let}, CodeBERT, CodeT5, and UniXCoder are each fine-tuned under two settings: (i) on the full PrimeVul train and val sets (175{,}797 training functions with 4{,}862 vulnerable and 170{,}935 benign; 23{,}948 validation functions), denoted as $\dagger$ in Table~\ref{tab:main} and reflecting the natural class-imbalanced distribution, and (ii) on the paired PrimeVul train and val subsets (3{,}789 vulnerable and 3{,}789 benign pairs), denoted as $\ddagger$ and matching the class-balanced distribution of the pair test set. For LineVul, we include two variants: the released checkpoint from its original study (denoted as $\S$) and its pair-subset fine-tuned counterpart ($\ddagger$). All DL models are built on top of the tokenizers and pretrained weights released by their respective authors.

For the LLM-based methods, we include four representative approaches: Ding et al.'s~\cite{ding2024primevul} Chain-of-Thought prompting (denoted as $*$), GPTLens, VulTrial, and ConColl. Ding et al.'s CoT is the best-performing prompting strategy reported in PrimeVul and guides the LLM to reason step by step before arriving at a decision. GPTLens~\cite{hu2023large} couples an auditor agent that proposes candidate vulnerabilities with a critic agent that scores them. VulTrial orchestrates four role-specific agents (security researcher, code author, moderator, and review board) under a courtroom-inspired setting and reaches its verdict through multi-round debate. ConColl~\cite{tsai2025sequential} follows a three-stage cascade composed of a single agent, a RAG-augmented agent, and a multi-agent collaboration, gated by confidence signals that terminate the cascade at the earliest stage yielding a high-certainty prediction. All LLM-based methods, including VulTriage, use GPT-4o~\cite{openai_gpt4o_2024} as the backbone. The detailed inference configuration is provided in Appendix~\ref{app:implementation_details}.

\noindent\textbf{Metrics.}
Following the evaluation protocol of PrimeVul and VulTrial, we adopt both pair-wise and standard classification metrics. The pair-wise metrics operate at the pair level and report which of the four joint outcomes occurs for each (vulnerable, benign) pair: \emph{Pair-wise Correct} (P-C) counts the pairs where both functions are classified correctly; \emph{Pair-wise Vulnerable} (P-V) and \emph{Pair-wise Benign} (P-B) count the pairs where both functions are labeled as vulnerable or benign, respectively; \emph{Pair-wise Reversed} (P-R) counts the pairs where the two labels are swapped; and \emph{Error} aggregates all pairs in which at least one function is misclassified. Among these metrics, P-C is our primary metric because it is the only outcome that simultaneously rules out false positives and false negatives within a pair, making it the most faithful indicator of fine-grained discrimination. In addition, treating vulnerable functions as the positive class, we report precision (P), recall (R), false positive rate (FPR), accuracy (Acc), and F1 score to provide a comprehensive view of classification behavior.

% ====================================================================
\subsection{Main Results}

Table~\ref{tab:main} reports the performance of VulTriage and all baselines on the PrimeVul pair test set. VulTriage achieves the best overall performance on the metrics that require fine-grained pair discrimination. Specifically, VulTriage attains 147 P-C, 288 Error, 0.6356 Accuracy, and 0.6907 F1. Compared with the strongest baseline in terms of P-C and Accuracy, VulTrial, VulTriage improves P-C by 66 pairs and Accuracy by 10.00 points. For F1, VulTriage exceeds the strongest LLM-based baseline, ConColl, by 6.20 points. Although some baselines obtain the best values on individual metrics such as R or FPR, these values mainly arise from degenerate label-biased predictions rather than stronger discrimination: for example, CodeBERT$^{\ddagger}$ and LineVul$^{\ddagger}$ almost always predict the vulnerable label, achieving very high recall but P-C = 0, while UniXCoder$^{\dagger}$ predicts most functions as benign, yielding the lowest FPR but only 4 P-C. In contrast, VulTriage achieves R = 0.8138 and FPR = 0.5425 while leading all baselines on P-C, Error, Accuracy, and F1, demonstrating a more balanced and reliable discriminative capability under the pair setting.

\begin{table}[t]
\centering
\caption{Ablation study of VulTriage components. $\uparrow$ indicates higher is better; $\downarrow$ indicates lower is better. Best results are in \textbf{bold}.}
\label{tab:ablation}
\renewcommand{\arraystretch}{1.15}
\setlength{\tabcolsep}{5pt}
\resizebox{0.95\columnwidth}{!}{%
\begin{tabular}{@{}lcccccccccc@{}}
\toprule
\textbf{Method}
 & \textbf{P-C$\uparrow$} & \textbf{P-V$\downarrow$} & \textbf{P-B$\downarrow$} & \textbf{P-R$\downarrow$} & \textbf{Error$\downarrow$}
 & \textbf{P$\uparrow$} & \textbf{R$\uparrow$} & \textbf{FPR$\downarrow$}
 & \textbf{Acc$\uparrow$} & \textbf{F1$\uparrow$} \\
\midrule
\rowcolor{gray!12}[\tabcolsep][\tabcolsep]
\textbf{VulTriage (Ours)}
 & \textbf{55} & \textbf{55} & 18 & \textbf{3} & \textbf{76}
 & \textbf{0.6548} & \textbf{0.8397} & \textbf{0.4427}
 & \textbf{0.6985} & \textbf{0.7358} \\
w/o Control Info
 & 42 & 58 & 23 & 8 & 89 & 0.6024 & 0.7634 & 0.5038 & 0.6298 & 0.6734 \\
w/o Knowledge
 & 45 & 63 & 13 & 10 & 86 & 0.5967 & 0.8244 & 0.5573 & 0.6336 & 0.6923 \\
w/o Explain
 & 42 & 60 & \textbf{11} & 18 & 89 & 0.5667 & 0.7786 & 0.5954 & 0.5916 & 0.6559 \\
\bottomrule
\end{tabular}%
}
\end{table}

\subsection{Ablation Study}

We design three ablation settings to validate the effectiveness of different components in VulTriage: (1) w/o Control Info, which removes the structural context generated from AST, CFG, and DFG; (2) w/o Knowledge, which removes the retrieved CWE-derived vulnerability knowledge; and (3) w/o Explain, which removes the semantic explanation produced before the final judgment. Following the 30\% evaluation budget, we report ablation results on 131 pairs from the PrimeVul pair test set. The overall results are reported in Table~\ref{tab:ablation}.

After removing each component, both pair-wise and standard classification metrics exhibit varying degrees of decline, which demonstrates the contribution of each path to the overall framework. Specifically, the full VulTriage achieves the best overall performance with 55 P-C, 76 Error, 0.6985 Accuracy, and 0.7358 F1. Removing the Control Path reduces P-C from 55 to 42 and increases Error from 76 to 89, showing that explicit structural information is important for distinguishing subtle vulnerable and benign code pairs. Removing the Knowledge Path also degrades the results, especially increasing FPR from 0.4427 to 0.5573, indicating that vulnerability-specific knowledge helps suppress false alarms. Notably, the w/o Explain setting shows the lowest Accuracy and F1 among all variants, suggesting that the semantic explanation is essential for helping the LLM understand the functional behavior of complex code. Overall, these results confirm that the three paths provide complementary information and jointly contribute to the effectiveness of VulTriage.

\subsection{Generalization to Low-Resource Programming Languages}

To further evaluate the applicability of VulTriage beyond the C/C++-dominated vulnerability detection setting, we conduct an additional study on a low-resource programming language, where Kotlin~\cite{le2024software} serves as the test case. Following ConColl, the Kotlin dataset consists of 20 vulnerable functions and 98 non-vulnerable functions. We use CodeQL~\cite{codeql} to extract language-specific structural information for the Control Path. Given the small data size and class imbalance, we adopt a 10-round evaluation strategy. In each round, the dataset is randomly split into 60\% for training, 20\% for validation, and 20\% for testing. The final metrics are averaged over the 10 rounds. To ensure a fair comparison with ConColl~\cite{tsai2025sequential}, which uses OpenAI's gpt-3.5-turbo as its underlying LLM, we also use GPT-3.5 as the backbone model for VulTriage in this experiment.

Table~\ref{tab:kotlin} reports the results on the Kotlin dataset. VulTriage achieves the best overall performance, obtaining 0.45 precision, 0.77 recall, and 0.57 F1-score. Compared with ConColl, VulTriage improves the F1-score from 0.44 to 0.57, while also improving both precision and recall. In particular, the recall improvement indicates that the proposed triple-path context augmentation helps identify more vulnerable functions under the low-resource and class-imbalanced setting. Meanwhile, VulTriage maintains the highest precision among all methods, suggesting that the improved recall does not come from trivially over-predicting vulnerabilities. These results demonstrate the promising generalization ability of VulTriage on low-resource programming languages.

\begin{table}[t]
\centering
\caption{Evaluation results on the Kotlin dataset. $\uparrow$ indicates higher is better. Best results are in \textbf{bold}.}
\label{tab:kotlin}
\renewcommand{\arraystretch}{1.15}
\setlength{\tabcolsep}{12pt}
\resizebox{0.5\columnwidth}{!}{%
\begin{tabular}{@{}lccc@{}}
\toprule
\textbf{Method}
 & \textbf{Pre$\uparrow$} & \textbf{Rec$\uparrow$} & \textbf{F1$\uparrow$} \\
\midrule
GPT Fine-Tuning
 & 0.37 & 0.32 & 0.34 \\
GPT Few-Shot
 & 0.43 & 0.44 & 0.43 \\
CodeBERT
 & 0.24 & 0.47 & 0.32 \\
ConColl
 & 0.31 & 0.73 & 0.44 \\
\textbf{VulTriage (Ours)}
 & \textbf{0.45} & \textbf{0.77} & \textbf{0.57} \\
\bottomrule
\end{tabular}%
}
\end{table}

\begin{figure*}[t]
\centering
\includegraphics[width=\textwidth]{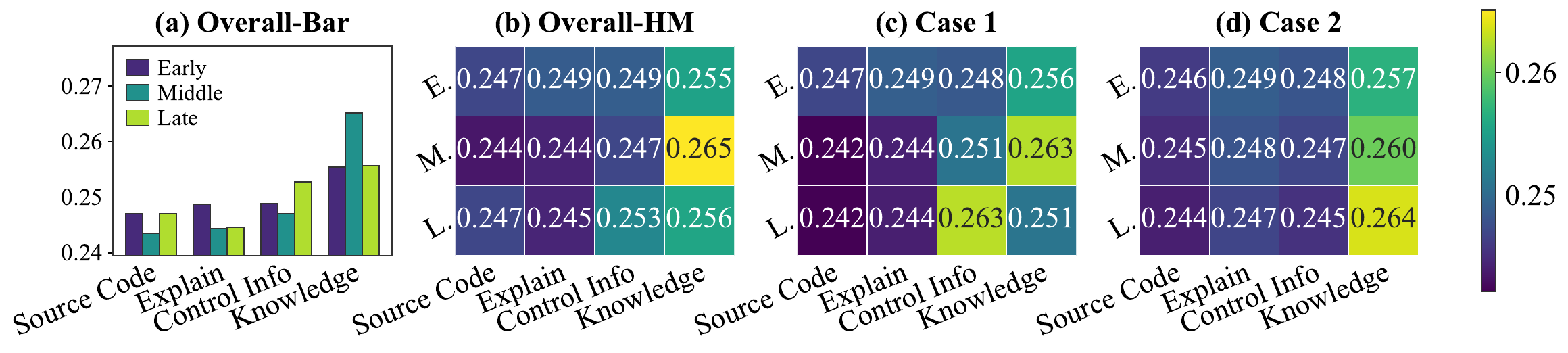}
\caption{Attention score analysis of different input blocks. E., M., and L. denote Early, Middle, and Late layers, respectively. (a) Overall bar statistics. (b) Overall heatmap. (c)--(d) Two representative cases.}
\label{fig:attention}
\end{figure*}

\subsection{Context Utilization Analysis}

To analyze how VulTriage utilizes different input contexts, we use Qwen3-8B~\cite{yang2025qwen3} as a transparent proxy model and compute its attention scores over four input blocks: Source Code, Control Info., Knowledge, and Explain. Since Qwen3-8B has 36 layers, we focus on the intermediate layers to reduce the influence of input- and output-adjacent representations. Specifically, we analyze layers 9--26 and group them into Early layers (9--14), Middle layers (15--20), and Late layers (21--26). We randomly sample 30\% of the test pairs for this analysis.

As shown in Figure~\ref{fig:attention}, all four input blocks receive non-trivial attention scores, indicating that the model uses multiple evidence sources rather than relying on a single context. Among them, Knowledge receives the highest overall attention, especially in the middle layers, suggesting that vulnerability-specific knowledge provides important domain anchors for the final judgment. Source Code and Explain are more emphasized in the early layers, showing that the model first attends to raw program content and functional interpretation. In contrast, Control Info. receives relatively higher attention in the late layers, implying that structural dependencies are more useful near the decision stage. These results indicate that the three augmented contexts provide complementary signals, helping VulTriage integrate code semantics, structural dependencies, and vulnerability priors.

\section{Conclusion}

We introduce VulTriage, an LLM-based vulnerability detection framework enhanced by Triple-Path Context Augmentation. VulTriage constructs three complementary contexts from structural analysis, vulnerability knowledge retrieval, and semantic understanding, enabling the LLM to better capture control/data dependencies, vulnerability-specific patterns, and functional program behavior. Experiments on the PrimeVul pair test set show that VulTriage outperforms both deep learning-based and LLM-based baselines, while ablation results confirm the effectiveness of each path. Additional results on the Kotlin dataset further demonstrate its generalization ability under low-resource and class-imbalanced settings. These findings suggest that structured, knowledge-aware, and semantic context augmentation is an effective direction for improving LLM-based vulnerability detection.

\clearpage

%%%%%%%%%%%%%%%%%%%%%%%%%%%%%%%%%%%%%%%%%%%%%%%%%%%%%%%%%%%%

%% ==== Required preamble packages ===============================
%% \usepackage[most]{tcolorbox}
%% \usepackage{fancyvrb}    % for Verbatim environment inside tcolorbox

\bibliography{dd}
\bibliographystyle{plainnat}

%%%%%%%%%%%%%%%%%%%%%%%%%%%%%%%%%%%%%%%%%%%%%%%%%%%%%%%%%%%%

%%%%%%%%%%%%%%%%%%%%%%%%%%%%%%%%%%%%%%%%%%%%%%%%%%%%%%%%%%%%

\appendix

\section{Limitations}
\label{app:lim}
In this work, we take the first step toward triple-path context augmentation for LLM-based vulnerability detection. While VulTriage achieves strong performance on PrimeVul and shows promising generalization on Kotlin, several limitations remain. First, the current experiments mainly focus on function-level vulnerability detection, and more fine-grained vulnerability localization can be further explored in future work. Second, our framework uses static structural information and CWE-derived knowledge as external contexts, while other useful security resources such as project-specific coding rules or historical patches are not yet incorporated. Third, the current prompt construction follows a fixed three-path design, and future work may investigate more adaptive context selection strategies for different vulnerability types and programming languages.

\section{Implementation Details}
\label{app:implementation_details}

To ensure reproducibility, we report the inference configuration used in our experiments in Table~\ref{tab:inference_params}. This configuration is used for LLM-based explanation generation and vulnerability prediction. In the Control Path, we use the Level-C granularity setting; in the Knowledge Path, we set the maximum number of retrieved CWE entries to 2. The average end-to-end running time per sample is 34.5 seconds.

\begin{table}[h]
\centering
\caption{Inference configuration used in our experiments.}
\label{tab:inference_params}
\renewcommand{\arraystretch}{1.15}
\begin{tabular}{lc}
\toprule
\textbf{Parameter} & \textbf{Value} \\
\midrule
Temperature & 0.7 \\
Top-p & 1.0 \\
Frequency penalty & 0 \\
Presence penalty & 0 \\
Timeout & 300 seconds \\
\bottomrule
\end{tabular}
\end{table}

\section{Statistical Significance Analysis}
\label{app:significance}

To further verify whether the improvements of VulTriage over its ablated variants are statistically reliable, we conduct paired McNemar's~\cite{mcnemar1947note} exact tests between the full model and each ablation setting. The results are reported in Table~\ref{tab:significance}. VulTriage shows statistically significant improvements over all variants. In particular, the comparison with w/o Explain reaches a stronger significance level of $p < 0.001$, indicating that the semantic explanation path contributes substantially to the overall performance. The improvements over w/o Control Info and w/o Knowledge are also significant with $p < 0.05$, further confirming the effectiveness of structural context and vulnerability knowledge. These results demonstrate that the performance gains of VulTriage are not incidental, but consistently supported by the three-path context augmentation design.

\begin{table}[t]
\centering
\caption{Statistical significance analysis between VulTriage and its ablated variants.}
\label{tab:significance}
\renewcommand{\arraystretch}{1.15}
\setlength{\tabcolsep}{10pt}
\resizebox{0.5\columnwidth}{!}{%
\begin{tabular}{@{}lc@{}}
\toprule
\textbf{Comparison} & \textbf{$p$-value} \\
\midrule
VulTriage vs. w/o Explain & $p < 0.001$ \\
VulTriage vs. w/o Control Info & $p < 0.05$ \\
VulTriage vs. w/o Knowledge & $p < 0.05$ \\
\bottomrule
\end{tabular}%
}
\end{table}

\section{Verbalization Templates of the Control Path}
\label{sec:appendix-templates}

This appendix details the verbalization operator $\tau$ used in Algorithm~\ref{alg:control}, which is the component that turns the three salient-view sets $\mathcal{V}_{\text{ast}}$, $\mathcal{V}_{\text{cfg}}$, and $\mathcal{V}_{\text{dfg}}$ into the natural-language fragments $T_{\text{ast}}$, $T_{\text{cfg}}$, and $T_{\text{dfg}}$. The operator is instantiated by a small set of pre-defined templates, each associated with a specific salient unit in one of the three views. A template contains a fixed natural-language skeleton and a set of placeholders (denoted in angle brackets, e.g., \texttt{<name>}), which are filled at runtime with values carried by the corresponding graph element. Table~\ref{tab:templates} lists all the templates used by $\tau$, grouped by the view they belong to.

\begin{table*}[h]
\centering
\small
\caption{Verbalization templates used by the shared operator $\tau$ in the Control Path. Placeholders in angle brackets are filled with values from the corresponding graph element at runtime.}
\label{tab:templates}
\renewcommand{\arraystretch}{1.35}
\resizebox{\textwidth}{!}{%
\begin{tabular}{@{}p{1.9cm} p{3.8cm} p{10.2cm}@{}}
\toprule
\textbf{View} & \textbf{Salient Unit} & \textbf{Template} \\
\midrule
\multirow{5}{*}{AST} 
 & Function summary  & \texttt{Function <name>@L<line>: <nd> declarations, <na> assignments, <nb> branches, <nc> calls.} \\
 & Key call chain    & \texttt{Key call chain: <fn$_1$>, <fn$_2$>, \ldots, <fn$_k$>.} \\
 & Conditions/Loops  & \texttt{Conditions/Loops: <stmt$_1$>; <stmt$_2$>; \ldots} \\
 & Returns           & \texttt{Returns: <expr$_1$>; <expr$_2$>; \ldots} \\
 & Isomorphic funcs  & \texttt{Isomorphic functions collapsed: <rep> represents <n> functions.} \\
\midrule
\multirow{3}{*}{CFG}
 & Function summary  & \texttt{Function <name>: retained control points <k>/<n>; branches <nb>; calls <nc>.} \\
 & Branch listing    & \texttt{Branch/Loop nodes: <label$_1$>@L<line$_1$>; <label$_2$>@L<line$_2$>; \ldots} \\
 & Skeleton path     & \texttt{Path <i>: Entry $\rightarrow$ [<edge-label>] <node$_1$> $\rightarrow$ [<edge-label>] <node$_2$> $\rightarrow$ \ldots $\rightarrow$ Exit.} \\
\midrule
\multirow{3}{*}{DFG}
 & Function summary  & \texttt{Function <name>: edges retained <k>/<n>; parameter sources <np>; chains <nc>.} \\
 & Parameter listing & \texttt{Parameter sources: <param$_1$>@L<line$_1$>; <param$_2$>@L<line$_2$>; \ldots} \\
 & Data chain        & \texttt{Data chain: <param> $\rightarrow$ <var$_1$> $\rightarrow$ \ldots $\rightarrow$ <sink>.} \\
\bottomrule
\end{tabular}%
}
\end{table*}

To illustrate how the templates are instantiated in practice, we walk through a concrete example on the C function \texttt{copy\_bytes} shown below, which copies a prefix of \texttt{src} into \texttt{dst} after performing a simple boundary check on the requested length \texttt{len}. The corresponding source code is given in Figure~\ref{box:src-code}, and the three natural-language fragments produced by $\tau$ are shown in Figures~\ref{box:t-ast}--\ref{box:t-dfg}.

\begin{figure}[h]
\centering
\begin{tcolorbox}[
    colback=codeColor!5,
    colframe=codeColor!70!black,
    title=\textbf{Source Code} ($C$),
    fonttitle=\bfseries,
    left=4pt, right=4pt, top=3pt, bottom=3pt,
    boxrule=0.6pt, arc=2pt
]
\begin{Verbatim}[fontsize=\small]
void copy_bytes(char *dst, const char *src, size_t len) {
    size_t max = MAX_BUF;
    if (len > max) {
        return;
    }
    memcpy(dst, src, len);
}
\end{Verbatim}
\end{tcolorbox}
\caption{Running-example source code of the \texttt{copy\_bytes} function.}
\label{box:src-code}
\end{figure}

\begin{figure}[h]
\centering
\begin{tcolorbox}[
    colback=ctrlColor!5,
    colframe=ctrlColor!70!black,
    title=\textbf{$T_{\text{ast}}$} --- AST fragment produced by $\tau$,
    fonttitle=\bfseries,
    left=4pt, right=4pt, top=3pt, bottom=3pt,
    boxrule=0.6pt, arc=2pt
]
\small
\texttt{Function copy\_bytes@L1: 2 declarations, 1 assignments, 1 branches, 1 calls. Key call chain: memcpy. Conditions/Loops: if(len > max).}
\end{tcolorbox}
\caption{AST fragment $T_{\text{ast}}$ produced by $\tau$ on the running example.}
\label{box:t-ast}
\end{figure}

\begin{figure}[h]
\centering
\begin{tcolorbox}[
    colback=ctrlColor!5,
    colframe=ctrlColor!70!black,
    title=\textbf{$T_{\text{cfg}}$} --- CFG fragment produced by $\tau$,
    fonttitle=\bfseries,
    left=4pt, right=4pt, top=3pt, bottom=3pt,
    boxrule=0.6pt, arc=2pt
]
\small
\texttt{Function copy\_bytes: retained control points 5/9; branches 1; calls 1. Path 1: Entry $\rightarrow$ [True] if(len > max) $\rightarrow$ return $\rightarrow$ Exit. Path 2: Entry $\rightarrow$ [False] if(len > max) $\rightarrow$ call memcpy $\rightarrow$ Exit.}
\end{tcolorbox}
\caption{CFG fragment $T_{\text{cfg}}$ produced by $\tau$ on the running example.}
\label{box:t-cfg}
\end{figure}

\begin{figure}[h]
\centering
\begin{tcolorbox}[
    colback=ctrlColor!5,
    colframe=ctrlColor!70!black,
    title=\textbf{$T_{\text{dfg}}$} --- DFG fragment produced by $\tau$,
    fonttitle=\bfseries,
    left=4pt, right=4pt, top=3pt, bottom=3pt,
    boxrule=0.6pt, arc=2pt
]
\small
\texttt{Function copy\_bytes: edges retained 3/4; parameter sources 2; chains 1. Parameter sources: param:len@L1; param:src@L1. Data chain: param:len $\rightarrow$ if(len > max) $\rightarrow$ call memcpy.}
\end{tcolorbox}
\caption{DFG fragment $T_{\text{dfg}}$ produced by $\tau$ on the running example.}
\label{box:t-dfg}
\end{figure}

Applying $\tau$ to the AST salient view fills the \emph{function summary}, \emph{key call chain}, and \emph{conditions/loops} templates with the values carried by \texttt{copy\_bytes}, yielding the fragment shown in Figure~\ref{box:t-ast}. For the CFG salient view, $\tau$ enumerates the two skeleton paths enabled by the boundary branch and renders them with the \emph{skeleton path} template, producing Figure~\ref{box:t-cfg}, in which the edge labels \texttt{[True]} and \texttt{[False]} reflect the two branches of \texttt{if(len > max)}. For the DFG salient view, $\tau$ instantiates the \emph{parameter listing} and \emph{data chain} templates, producing Figure~\ref{box:t-dfg} that makes explicit how the parameter \texttt{len} reaches the \texttt{memcpy} call through the boundary check. The three fragments are concatenated in the order AST, CFG, DFG to form the final structural context $\mathbf{S} = T_{\text{ast}} \oplus T_{\text{cfg}} \oplus T_{\text{dfg}}$ that is injected into the instruction $\mathcal{T}$.

\section{Broader Impacts}
\label{app:broader_impacts}

This work has potential positive societal impacts by improving the reliability and accessibility of automated software vulnerability detection. Software vulnerabilities remain a major threat to modern digital infrastructure, and inaccurate detection tools may either miss security-critical bugs or overwhelm developers with false alarms. By augmenting LLM-based vulnerability detection with structural program information, CWE-derived vulnerability knowledge, and semantic explanations, VulTriage aims to provide more reliable evidence for vulnerability reasoning and help developers identify potentially vulnerable code more effectively. Such tools can assist security practitioners in prioritizing manual audits, reduce the burden of reviewing large codebases, and support earlier detection of vulnerabilities during software development.

At the same time, VulTriage may also introduce potential risks if used improperly. False negatives may give developers a misleading sense of security, while false positives may waste auditing effort or distract developers from real vulnerabilities. Developers may also over-rely on the model output and skip necessary manual verification. In addition, more accurate vulnerability localization could be misused by attackers to identify exploitable code regions. Therefore, VulTriage should be used as an assistive auditing tool rather than a replacement for expert security review, and its outputs should be verified before deployment or disclosure.

In addition, VulTriage may benefit resource-constrained security scenarios. Since the framework relies on context augmentation rather than task-specific model training, it can be applied more flexibly to settings where labeled vulnerability data are limited. Our additional evaluation on Kotlin suggests that this direction may help extend vulnerability detection support to low-resource programming languages, where dedicated security tools and labeled datasets are often less mature. More broadly, improving the accuracy and interpretability of vulnerability detection systems can contribute to safer software ecosystems and help developers build more secure applications.

% Required packages:
% \usepackage{xcolor}
% \usepackage{tcolorbox}
% \tcbuselibrary{breakable,skins,listings}

\definecolor{PromptBack}{RGB}{248,248,248}
\definecolor{PromptFrame}{RGB}{210,210,210}

\newtcblisting{promptbox}[2][]{
  enhanced,
  breakable,
  colback=PromptBack,
  colframe=PromptFrame,
  boxrule=0.6pt,
  arc=1.5mm,
  left=1.5mm,
  right=1.5mm,
  top=1mm,
  bottom=1mm,
  title={#2},
  fonttitle=\bfseries,
  listing only,
  listing options={
    basicstyle=\ttfamily\small,
    breaklines=true,
    columns=fullflexible,
    keepspaces=true,
    showstringspaces=false
  },
  #1
}

\section{Prompt Templates}
\label{app:prompt_templates}

This section provides the prompt templates used in VulTriage. Each template corresponds to one LLM-based agent call in the framework. The placeholders enclosed by angle brackets are replaced with the corresponding input content during inference.

\subsection{Knowledge Path: Vulnerability Query Generation}
\label{app:prompt_knowledge_query}

The Knowledge Path first asks the LLM to produce coarse vulnerability-oriented natural-language queries from the input code. These queries are then used to retrieve relevant CWE-derived vulnerability descriptions and examples.

\begin{promptbox}{Prompt Template for Vulnerability Query Generation}
You are an expert software security analyst.

Given the following source code, identify at most two possible vulnerability types or weakness patterns that may be relevant to this code. Your output will be used as retrieval queries for a vulnerability knowledge base, so each query should be concise, security-specific, and written in natural language.

If the code appears unlikely to contain any vulnerability, output a generic query that can still help retrieve common vulnerability inspection knowledge.

Source Code:
<Code>

Output format:
Query 1: <a concise vulnerability description>
Query 2: <a concise vulnerability description or N/A>
\end{promptbox}

\subsection{Semantic Path: Functional Explanation Generation}
\label{app:prompt_semantic_explanation}

The Semantic Path asks the LLM to summarize the observable functional behavior of the input code. This explanation provides a denoised semantic view before the final vulnerability judgment.

\begin{promptbox}{Prompt Template for Semantic Explanation Generation}
You are an expert program analysis assistant.

Given the following source code, summarize its observable functional behavior. Focus on what the code directly does rather than speculating about hidden intent.

Please describe:
1. The main purpose of the function.
2. Important inputs, parameters, and return values.
3. Key operations, including memory operations, pointer operations, file operations, system calls, arithmetic operations, and boundary checks if present.
4. Observable data flow from inputs to sensitive operations when applicable.
5. Any security-relevant behavior that can be inferred directly from the code.

Do not decide whether the code is vulnerable in this step. Only provide a concise functional explanation.

Source Code:
<Code>

Functional Explanation:
\end{promptbox}

\subsection{Final Judgment Agent: Vulnerability Prediction}
\label{app:prompt_final_judgment}

The final judgment agent receives the original source code together with the three augmented contexts: structural context from the Control Path, CWE-derived knowledge from the Knowledge Path, and semantic explanation from the Semantic Path. It then predicts whether the function is vulnerable.

\begin{promptbox}{Prompt Template for Final Vulnerability Judgment}
You are an expert software security engineer.

Your task is to determine whether the given source code contains a potential vulnerability. You are provided with four types of information:

1. Source Code:
<Code>

2. Control Information:
<Control_Info>

3. Vulnerability Knowledge:
<Knowledge>

4. Functional Explanation:
<Explain>

Please analyze the code by considering:
- whether user-controlled or external data can reach sensitive operations;
- whether control-flow and data-flow dependencies indicate missing checks;
- whether the retrieved vulnerability knowledge matches the observed code behavior;
- whether the functional explanation reveals risky memory, pointer, file, arithmetic, or system-level operations;
- whether there are sufficient guards, validations, or boundary checks.

Return the final prediction using the following strict format:

Verdict: <Yes or No>

\end{promptbox}

\end{document}